\pdfoutput=1

\documentclass[11pt]{article}

\usepackage[final]{coling}

\usepackage{times}
\usepackage{latexsym}
\usepackage[T1]{fontenc}

\usepackage[utf8]{inputenc}

\usepackage{microtype}

\usepackage{inconsolata}

\usepackage{graphicx}
\usepackage{xcolor}
\usepackage{booktabs}
\usepackage{mdframed}
\usepackage{inconsolata}
\usepackage{multirow}
\usepackage{amsmath}
\usepackage{tcolorbox}
\usepackage{amssymb}
\usepackage{bbding}
\usepackage[normalem]{ulem}
\useunder{\uline}{\ul}{}
\newcommand{\bench}{\textsc{WikiGenBench}}
\title{\bench{}: Exploring  Full-length Wikipedia Generation \\under Real-World Scenario}

\author{Jiebin Zhang \thanks{$^*$Equal contribution. }$^{*\dag}$, Eugene J. Yu{$^{**\dag}$}, Qinyu Chen{$^\dag$}, 
Chenhao Xiong$^\dag$ Dawei Zhu$^\dag$\\
\textbf{
Han Qian$^\dag$,
Mingbo Song$^\dag$, Weimin Xiong$^\dag$
Xiaoguang Li$^\ddag$}, \textbf{Qun Liu$^\ddag$}, \textbf{Sujian Li$^\dag$} \\
  $^\dag$ National Key Laboratory for Multimedia Information Processing, \\School of Computer Science, Peking University \\
  \texttt{\{zhangjiebin,ejyu,chenqinyu,dwzhu,lisujian\}@pku.edu.cn} \\
  $^\ddag$ Huawei Technologies\\
  \texttt{\{lixiaoguang11,qun.liu\}@huawei.com} 
  }

\begin{document}
\maketitle
\begin{abstract}
It presents significant challenges to generate comprehensive and accurate Wikipedia articles for newly emerging events under a real-world scenario. 
Existing attempts fall short either by focusing only on short snippets or by using metrics that are insufficient to evaluate real-world scenarios.
In this paper, we construct \bench{}, a new benchmark consisting of 1,320 entries, designed to align with real-world scenarios in both generation and evaluation. For generation, we explore a real-world scenario where structured, full-length Wikipedia articles with citations are generated for new events using input documents from web sources.  For evaluation, we integrate systematic metrics and LLM-based metrics to assess the verifiability, organization, and other aspects aligned with real-world scenarios. Based on this benchmark, we conduct extensive experiments using various models within three commonly used frameworks: direct RAG, hierarchical structure-based RAG, and RAG with a fine-tuned generation model.  Experimental results show that hierarchical-based methods can generate more comprehensive content, while fine-tuned methods achieve better verifiability. However, even the best methods still show a significant gap compared to existing Wikipedia content, indicating that further research is necessary.\footnote{The data and code can be accessed at \url{https://github.com/zhzihao/WikiGenBench}.}
\end{abstract}

\section{Introduction}

Wikipedia serves as an indispensable repository for high-quality summaries encompassing a broad spectrum of subjects~\citep{lemmerich2019world}. Its rich informativeness and reliability make it an invaluable asset for numerous knowledge-intensive NLP tasks, such as information retrieval~\citep{lehmann2015dbpedia,sharma2024exploring}, question answering~\citep{chen2017reading,yang2018hotpotqa}, and automatic summarization~\citep{liu2018generating}.  However, existing practices in constructing Wikipedia heavily rely on human curation, which struggles to keep pace with the exponential growth of new events and subjects across the Internet~\citep{2019t5,biderman2022datasheet}. Consequently, the automatic generation of high-quality Wikipedia articles has become an urgent need.

Many efforts have been devoted to generating Wikipedia articles, yet they still fall short of real-world applicability. A large portion of earlier work~\citep{sauper2009automatically,liu2018generating,perez2019generating,banerjee2016wikiwrite} focused on generating short Wikipedia snippets, such as the first paragraph of an article, rather than full-length entries. This approach does not align with the complexity of real-world Wikipedia generation, which requires longer, well-structured articles with proper citations. Recent efforts~\citep{fan2022generating,qian2023webbrain} emphasize article structure and citation inclusion, but still heavily rely on traditional metrics like ROUGE~\citep{lin-2004-rouge} and QG-QA~\citep{goodrich2019assessing,wang2020asking}, which are insufficient for evaluating Wikipedia generation in real-world scenarios. 

\begin{figure}[t]
    \centering
    \includegraphics[width=0.85\linewidth]{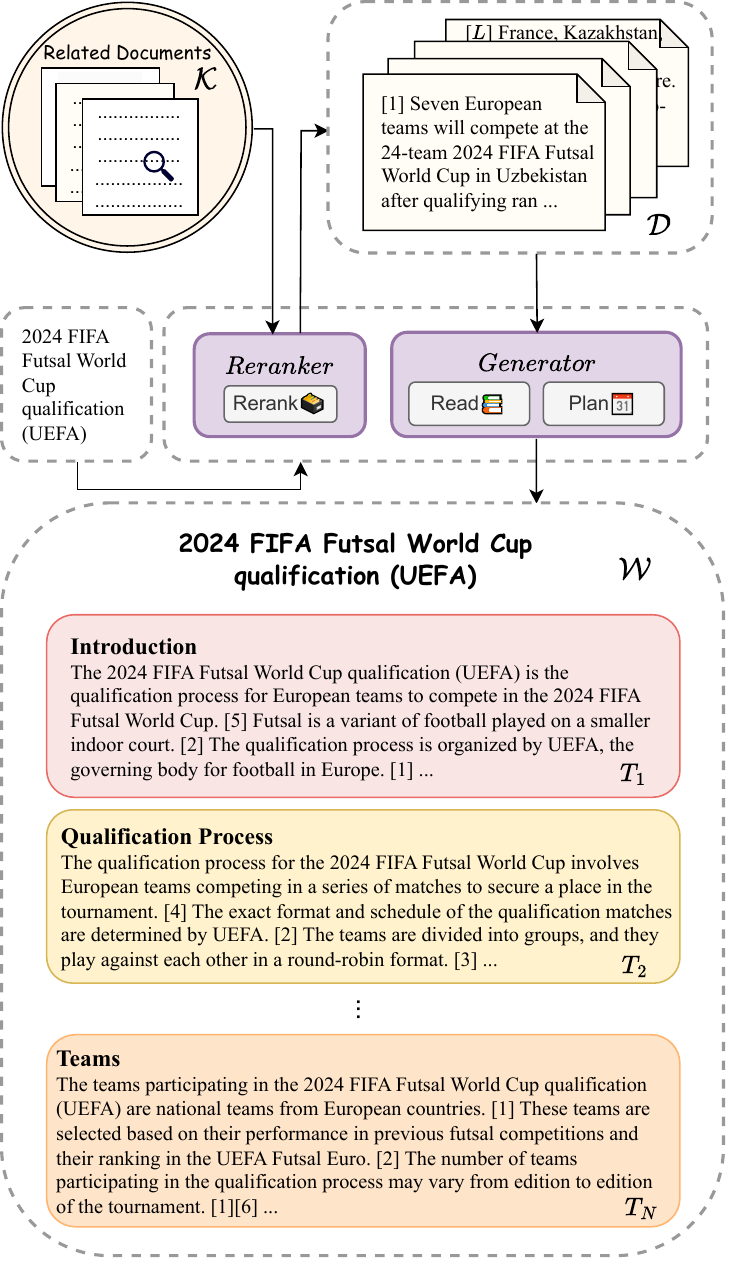}
 \caption{Illustration of the proposed Wikipedia generation task.}
 \label{intro_model}
\end{figure}

In this paper, we ensure that both the task definition and the benchmark evaluation align with real-world scenarios, as shown in Figure \ref{intro_model}.
We reformulate the task as generating structured, full-length Wikipedia articles with citations. To support this, we construct a dataset of 1,320 new events, along with 
reference documents sourced from the Internet. We take care to minimize model pre-exposure by focusing primarily on events that occurred after the knowledge cutoff date of our main experimental models~\citep{ouyang2022training, touvron2023llama}. This helps mitigate pre-exposure effects, ensuring that generation relies primarily  on the provided reference documents.
At the same time, we assess the task across three key dimensions: writing, informativeness, and verifiability following Wikipedia's evaluation standards. We break down the assessment into more granular aspects and use LLM-based methods~\citep{achiam2023gpt,kim2024prometheus} for the evaluation of writing and informativeness, which have proven highly effective~\citep{sottana2023evaluation,chiang-lee-2023-large,lin2023llmeval}. We employ the method from ALCE~\citep{gao2023enabling} to better assess the verifiability, which is a crucial criterion in Wikipedia\footnote{\url{https://en.wikipedia.org/wiki/Wikipedia:Verifiability}}.

To investigate LLM capabilities in Wikipedia generation, we develop baseline methods under the Retrieval-Augmented Generation (RAG) framework \citep{lewis2020retrieval, izacard2020leveraging, hu2023survey}. Our goal is to use state-of-the-art RAG techniques to retrieve important information for Wikipedia generation \cite{gao2023retrieval, ma2023query, shao2023enhancing}. With this aim, we develop three frameworks: a naive RAG approach called Retrieve-then-Read (\textbf{RR}), an advanced RAG method called Plan-Retrieve-Read (\textbf{PRR}), and a finetuned RR model (\textbf{TunedRR}). RR reranks related documents and reads the top ones for generation, while PRR uses LLMs' planning capabilities and a multi-stage reranking strategy to outline and generate articles section by section. TunedRR employs a fine-tuning strategy for Wikipedia generation. We compare the performance of multiple LLM models across these three frameworks.
We also examine the impact of different retrieval techniques and citation sources under the RR setting.

Experimental results show that hierarchical-based methods can produce more comprehensive content, while fine-tuned methods achieve better verifiability. Open-source models still lag significantly behind proprietary models in terms of verifiability.
We also observe, even with the best methods, the generated content falls short in average quality compared to original Wikipedia articles.
Our work provides the first systematic comparison of LLM-based methodologies for full-length Wikipedia generation under real-world scenarios, offering valuable insights and highlighting the potential of combining retrieval techniques with LLM models to improve Wikipedia generation quality.

\section{\bench{}}
\label{sec:benchmark}

\subsection{Task Formalization}
\label{task definition}

In this section, we formally define the task of Wikipedia generation. Given an event and its related documents $\mathcal{K}$, pre-collected through  search engines or manually extracted by human editors from the internet, this task aims to generate a full-length Wikipedia article $\mathcal{W}$ with $N$ sections $\{T_1, T_2, ..., T_N\}$ and $M$ sentences $\{S_1, S_2, ..., S_M\}$, as in Figure \ref{intro_model}.

\begin{normalsize}
\small{
\begin{align}
    \mathcal{D} = \{D_1, ..., D_L\} = \textit{Reranker}(\mathcal{K})
\end{align}
\begin{align}
    \mathcal{W} = T_1 \cup T_2 \cup \ldots \cup T_N = \textit{Generator}(\mathcal{D})
\end{align}
\begin{align}
    T_i &= \{<S_1, C_1>, \ldots, <S_{M_i}, C_{M_i}>\} \notag \\[2pt] 
    &\text{ } \qquad |\mathcal{W}| = M = \sum_{i=1}^{N} M_i 
\end{align}
}
\end{normalsize}

$T_i$  denotes the $i^{th}$ section and is composed of $M_i$ sentences. Each generated sentence $S_j$ belongs to a specific section of the article (e.g., $T_i$), maintaining a clear structure. To ensure verifiability, every sentence $S_j$ is accompanied by its corresponding citations $C_j \subset \mathcal{K}$. This definition ensures the generated Wikipedia article is coherent, well-organized, and substantiated with verifiable sources.

To generate a well-written and structured full-length Wikipedia article, the process usually involves two main components: a \textit{reranker} and a \textit{generator}. The LLM-based generator, either directly invoked or fine-tuned, reads the reranked set of documents $\mathcal{D}$ and generates the article. The reranker reorders the documents according to the generator's requirements, prioritizing the most relevant information.

\subsection{Dataset Construction}
To achieve our task, we collect Wikipedia entries, including articles, section outlines, and related documents from human editors. During dataset construction, we select the most recent Wikipedia entries of events to mitigate potential training data leakage. Other types of entries like celebrities, which have extensive historical records, are excluded to avoid being seen by LLMs. Additionally, we maintain entries with word counts between 1000 and 3000 to align with typical Wikipedia article lengths and filter out low-resource entries that may not serve as good evaluation cases. This approach helps ensure the dataset's relevance and quality.

Due to the high failure rate of Wikipedia’s internal reference links, we also utilize Google's search API to retrieve relevant web pages based on article titles, obtaining related documents from both human editors and search engines for comparison. We focus on Wikipedia entries about events, creating a dataset of 1,320 entries. Using the latest events minimizes the likelihood that the model has been trained on related Wikipedia data, which is crucial for evaluating the model's ability to present factual information. We select 309 entries as the test set and use the remaining entries for training (Appendix \ref{appendix:training}).
\subsection{Dataset Statistics}
\begin{table}[t]
\centering
\resizebox{\columnwidth}{!}{
\begin{tabular}{@{}c|lr@{}}
\toprule
Source & Dataset Statistics & \multicolumn{1}{c}{} \\ \midrule
\multirow{2}{*}{Reference Wikipedia} & Sections (avg.) & 5.97 \\
 & Word count (avg.) & 1665.51 \\ \midrule
\multirow{2}{*}{\begin{tabular}[c]{@{}c@{}}Related documents by\\ human editor\end{tabular}} & \# Related docs (avg.) & 17.49 \\
 & Word count (avg.) & 13k \\ \midrule
\multirow{2}{*}{\begin{tabular}[c]{@{}c@{}}Related documents by\\ search engine\end{tabular}} & \# Related docs (avg.) & 24.12 \\
 & Word count (avg.) & 24k \\ \midrule
\multirow{3}{*}{\begin{tabular}[c]{@{}c@{}}Reference Wikipedia\\ +\\  Related documents\end{tabular}} & Events & 1320 \\
 & \# Related docs & 55k \\
 & Word count & 107M \\ \bottomrule
\end{tabular}
}
\caption{Statistics of \bench{} dataset. We report the scale of Wikipedia reference articles and related documents.}
\label{tab:dataset_characteristics_overview}
\end{table}

Table \ref{tab:dataset_characteristics_overview} outlines the statistics of the \bench{} dataset, which totally consists of 107 million words in total across 1,320 Wikipedia entries and 55k related documents. Wikipedia articles in the dataset have an average of 1,665 words and 5.97 sections. Related documents curated by humans average 13k words, whereas documents retrieved via search engine average 24k words. 
These related documents substantially cover the information and knowledge about the events, making them suitable input for the automatic generation of Wikipedia articles. 
More details about the dataset and the collection process can be found in Appendix \ref{wikipages}.

\section{Evaluation Metrics}
We assess the task according to Wikipedia's evaluation criteria\footnote{\url{https://en.wikipedia.org/wiki/Wikipedia:Assessing\_articles}}, focusing on three dimensions: writing, informativeness, and verifiability. Recent studies \citep{sottana2023evaluation,chiang-lee-2023-large,lin2023llmeval} have demonstrated the effectiveness of using LLMs as evaluators. In our evaluation, we design LLM-based metrics with appropriate prompts, specifically utilizing GPT4~\cite{achiam2023gpt} and Prometheus2 \cite{kim2024prometheus}, a 7B LLM-based evaluator. Metrics are rated on a 0-5 scale, with detailed prompts provided in Appendix \ref{evaluation-appendix}.
In writing, we incorporate fluency and extend to measure outlining and organization. For informativeness,
we measure content coverage and focus, and include n-gram-based metrics like ROUGE, METEOR. 
Verifiability primarily measures whether a sentence  $S_j$ is supported by its citation  $C_j$, which is a crucial criterion in Wikipedia. We utilize the metrics from ALCE~\cite{gao2023enabling} to better assess this aspect.
Compared to previous work (Appendix \ref{appendix:evaluation}), we have developed a more comprehensive set of automatic evaluation metrics. Subsequently, we will provide a detailed explanation of the metrics used for each dimension.

\noindent \textbf{Writing}\quad 
We design three metrics to evaluate Wikipedia articles: \textit{Fluency}, \textit{Organization}, and \textit{Outline Scores}. The \textit{Fluency Score} assesses fluency and readability, the \textit{Organization Score} evaluates structure and logical connections, and the \textit{Outline Score} checks section heading quality. We utilize Prometheus2 to assess the \textit{Organization Score} \cite{shao2024assisting}, and GPT4 to assess the \textit{Fluency Score} and \textit{Outline Score}.

\noindent \textbf{Informativeness}\quad 
We compile n-gram-based metrics including METEOR~\citep{banerjee-lavie-2005-meteor}, ROUGE-L~\citep{lin-2004-rouge}, which are widely used in text generation. Higher values in these metrics indicate greater similarity between the generated and reference texts. These metrics are computed using the NLG-eval package~\citep{sharma2017relevance}. Since the Wikipedia reference article may not always be the "gold answer," we design the \textit{Info Score} to evaluate the overall richness of the generated content. The \textit{Focus Score} examines whether the article remains on topic and maintains a clear focus, while the \textit{Coverage Score} determines if the article provides an in-depth exploration of the topic. We use Prometheus2 to evaluate both \textit{Focus Score} and \textit{Coverage Score}~\cite{shao2024assisting}.

\noindent \textbf{Verifiability}\quad 
Verifiability measures whether the information in a Wikipedia article comes from a reliable source\footnote{\url{https://en.wikipedia.org/wiki/Wikipedia:Verifiability}}. To assess our model's verifiability, we develop three metrics: Citation Recall, Citation Precision~\citep{liu-etal-2023-evaluating}, and Citation Rate. 
To implement the measures,  
we utilize the NLI model TRUE~\citep{honovich-etal-2022-true}. 
We define $\phi(C_{i,j}, S_i) = 1$ if the citation $C_{i,j}$ entails the sentence $S_i$. For sentence $S_i$ and its corresponding citations $C_i = \{C_{i,1}, ..., C_{i,O_i}\}$:

\begingroup
\small
\begin{align}
&\text{}\text{Citation Recall} = \frac{1}{M} \sum_{i=1}^{M} \mathbb{I} \left( \max_{j} \phi(C_{i,j}, S_i) = 1 \right) \\
&\text{}\text{Citation Precision} = \frac{1}{M} \sum_{i=1}^{M} \left( \frac{\sum_{j=1}^{O_i} \phi(C_{i,j}, S_i)}{O_i} \right) \\
&\text{}\text{Citation Rate} = \frac{\sum_{i=1}^{N} (\text{\#words}(S_i) \cdot \text{Citation Recall}_i)}{\text{\#words}(\mathcal{W})}
\end{align}
\endgroup

As shown in the equations, Citation Recall is the proportion of sentences with at least one valid citation, where $\mathbb{I}$ denotes the indicator function, returning 1 if a condition is true and 0 if false. Citation Precision is the average proportion of valid citations per sentence. To rectify the influence of sentence length, Citation Rate is the weighted average of each sentence's Citation Recall, with the weights being the number of words in the sentences. 

\section{Baseline Methods}
According to task description in Section \ref{task definition}, we design the following three types of generation frameworks: 


\noindent \textbf{RR (Retrieve-then-Read)}\quad RR is a naive RAG framework. We follow the "Retrieve-then-Read" method from \citet{ma2023query} and adapt it for Wikipedia article generation. In this framework, a reranker orders reference documents based on their relevance to the event keyword and provides the top $L$ documents to the generator. The generator, a frozen LLM, reads these documents and directly generates the Wikipedia article.

\noindent \textbf{PRR (Plan-Retrieve-Read)}\quad
PRR is an advanced RAG framework inspired by hierarchical generation techniques in long story and dialogue generation~\cite{fan-etal-2018-hierarchical, bansal-etal-2022-r3}. PRR first uses a frozen LLM to plan the overall structure and generate section headings based on the reference documents. For each section, PRR employs the "Retrieve-then-Read" strategy to rerank related documents according to section headings and event keywords, and then generates the content for each section. We then aggregate the content of each section together to form the final Wikipedia article. 

\noindent \textbf{TunedRR}\quad 
The method is inspired by the idea that a small amount of data can teach a model to follow instructions, as shown in LIMA~\citep{zhou2023lima}. TunedRR aims to finetune the generator based on the RR model. This requires a training dataset of input related documents and corresponding output Wikipedia articles. While the related documents from our evaluation data can be used, the associated Wikipedia articles cannot due to the high failure rate of citation links~\citep{liu2018generating}. To address this, we leverage the robust performance of GPT4. We feed the related documents into GPT4 and use the same prompt as RR to generate synthesized Wikipedia articles. This process produced 1,011 data samples, which we used to train Llama2 and Vicuna models.

\section{Experiments}
\subsection{Experimental Settings}
Our baseline methods include proprietary ChatGPT variants (GPT3.5-turbo and GPT3.5-turbo-16k) and open-source LLMs (Llama2-7b-chat, Llama2-13b-chat \citep{touvron2023llama}, Vicuna-7b-v1.5, and Vicuna-13b-v1.5 \citep{vicuna2023}). FastChat is employed to enhance inference efficiency in open-source LLMs. All models use the same prompt (details in Appendix \ref{Prompt design}) to ensure evaluation fairness. For rerankers, we use traditional sparse word-based techniques like TF-IDF \cite{ramos2003using} and BM25 \cite{robertson2004simple}, as well as advanced dense vector-based retrievers, including DPR \citep{karpukhin-etal-2020-dense} and GTR \citep{ni-etal-2022-large}.

We segment related documents into 256-word chunks~\citep{borgeaud2022improving} before feeding them into the reranker module to reduce the computational burden of verifiability evaluation~\citep{gao2023enabling}. The reranker uses DPR by default and selects the top 5 chunks. For TunedRR, we utilize FastChat to employ full parameter fine-tuning with default hyperparameters. 
Additionally, we evaluate human-authored Wikipedia articles, as shown in the first row of Table \ref{main-results}. We do not assess the quality of their citations due to the numerous physical documents and links that are difficult to crawl.

\begin{table*}[h]
\renewcommand{\arraystretch}{1.1}
\centering
\resizebox{\linewidth}{!}{
\begin{tabular}{@{}lcccccccccccc@{}}
\toprule
\multicolumn{1}{c|}{\multirow{2}{*}{\textbf{Models}}} & \multicolumn{3}{c|}{\textbf{Writing}} & \multicolumn{5}{c|}{\textbf{Informativeness}} & \multicolumn{3}{c|}{\textbf{Verifiability}} & \multirow{2}{*}{\textbf{Length}} \\
\multicolumn{1}{c|}{} & \textit{\begin{tabular}[c]{@{}c@{}}Fluency \\ Score\end{tabular}} & \textit{\begin{tabular}[c]{@{}c@{}}\textit{Org.} \\ \textit{Score}\end{tabular}}& \multicolumn{1}{c|}{\textit{\begin{tabular}[c]{@{}c@{}}Outline \\ Score\end{tabular}}}  & MET & R-L & \textit{\begin{tabular}[c]{@{}c@{}}Focus \\ Score\end{tabular}}  & \textit{\begin{tabular}[c]{@{}c@{}}\textit{Cover.} \\ Score\end{tabular}}& \multicolumn{1}{c|}{\textit{\begin{tabular}[c]{@{}c@{}}Info\\ Score\end{tabular}}} & \begin{tabular}[c]{@{}c@{}}Cit. \\ Rate\end{tabular} & \begin{tabular}[c]{@{}c@{}}Cit. \\ Recall\end{tabular} & \multicolumn{1}{c|}{\begin{tabular}[c]{@{}c@{}}Cit. \\ Prec.\end{tabular}} &  \\ \midrule
\multicolumn{1}{l|}{Reference Wikipedia} & \multicolumn{1}{c}{4.45} & 3.61 & \multicolumn{1}{c|}{2.64} & \textbf{-} & \textbf{-} & 4.02 & 4.10 & \multicolumn{1}{c|}{4.83} & \textbf{-} & \textbf{-} & \multicolumn{1}{c|}{\textbf{-}} & 1639.0 \\ \midrule
\multicolumn{13}{c}{\textit{RR (Retrieve-then-Read)}} \\ \midrule
\multicolumn{1}{l|}{GPT3.5-turbo-0613} & \multicolumn{1}{c}{\textbf{4.31}} & \underline{4.05} & \multicolumn{1}{c|}{\textbf{2.86}} & \textbf{10.73} & \textbf{17.81} & \underline{4.26}& \textbf{3.94}& \multicolumn{1}{c|}{\textbf{3.49}} & \textbf{42.09} & \textbf{38.78} & \multicolumn{1}{c|}{\textbf{36.70}} & 579.1 \\
\multicolumn{1}{l|}{GPT3.5-turbo-1106-16k} & \multicolumn{1}{c}{4.29} & 4.02 & \multicolumn{1}{c|}{\underline{2.84}} & \underline{10.29} & \underline{17.42} & 4.22 & 3.89 & \multicolumn{1}{c|}{\underline{3.39}} & \underline{38.38} & \underline{33.98} & \multicolumn{1}{c|}{\underline{32.68}} & 541.3 \\
\multicolumn{1}{l|}{Llama2-7b} & \multicolumn{1}{c}{3.87} & 3.64 & \multicolumn{1}{c|}{1.43} & 10.21 & 16.05 & 3.77 & 3.27 & \multicolumn{1}{c|}{2.94} & 10.16 & 15.85 & \multicolumn{1}{c|}{15.83} & 625.7 \\
\multicolumn{1}{l|}{Llama2-13b} & \multicolumn{1}{c}{3.97} & \textbf{4.16}& \multicolumn{1}{c|}{2.39} & 9.74 & 15.89 & \textbf{4.38} & \underline{3.91} & \multicolumn{1}{c|}{3.03} & 7.91 & 8.91 & \multicolumn{1}{c|}{8.91} & 552.9 \\
\multicolumn{1}{l|}{Vicuna-7b} & \multicolumn{1}{c}{4.06} & 3.46 & \multicolumn{1}{c|}{1.61} & 10.18 & 17.34 & 3.69 & 3.39 & \multicolumn{1}{c|}{3.27} & 6.40 & 4.41 & \multicolumn{1}{c|}{4.38} & 535.2 \\ 
\multicolumn{1}{l|}{Vicuna-13b} & \multicolumn{1}{c}{\underline{4.18}} & 3.72 & \multicolumn{1}{c|}{2.27} & 9.80 & 17.33 & 3.98 & 3.63 & \multicolumn{1}{c|}{\underline{3.39}} & 16.88 & 11.03 & \multicolumn{1}{c|}{10.70} & 491.8 \\ \midrule 
\multicolumn{13}{c}{\textit{PRR (Plan-Retrieval-Read)}} \\ \midrule
\multicolumn{1}{l|}{GPT3.5-turbo-0613} & \multicolumn{1}{c}{\textbf{4.02}} & 3.36 & \multicolumn{1}{c|}{2.76} & 22.29 & 22.26 & \underline{3.69} & \textbf{3.51} & \multicolumn{1}{c|}{\textbf{3.78}} & \textbf{50.96} & \underline{53.47} & \multicolumn{1}{c|}{\underline{52.43}} & 1991.2 \\
\multicolumn{1}{l|}{GPT3.5-turbo-1106-16k} & \multicolumn{1}{c}{\textbf{4.02}} & \underline{3.38}& \multicolumn{1}{c|}{2.76} & 22.24& \underline{22.27}& \textbf{3.70} & \textbf{3.51} & \multicolumn{1}{c|}{\underline{3.76}} & 50.6 & \textbf{53.57} & \multicolumn{1}{c|}{\textbf{52.53}} & 1988.9 \\
\multicolumn{1}{l|}{Llama2-7b} & \multicolumn{1}{c}{2.83} & 2.74 & \multicolumn{1}{c|}{2.52} & 24.14 & 21.16 & 2.87 & 2.69 & \multicolumn{1}{c|}{2.27} & 13.08 & 27.02 & \multicolumn{1}{c|}{26.89} & 4210.4 \\
\multicolumn{1}{l|}{Llama2-13b} & \multicolumn{1}{c}{\underline{3.70}} & \textbf{3.41} & \multicolumn{1}{c|}{2.70} & \underline{24.18} & 20.94 & 3.68 & \underline{3.40} & \multicolumn{1}{c|}{3.29} & 12.77 & 14.30 & \multicolumn{1}{c|}{14.30} & 3789.5 \\
\multicolumn{1}{l|}{Vicuna-7b} & \multicolumn{1}{c}{2.89} & 2.40 & \multicolumn{1}{c|}{\underline{2.71}} & 23.43 & 21.87 & 2.61 & 2.68 & \multicolumn{1}{c|}{2.61} & 14.89 & 22.40 & \multicolumn{1}{c|}{22.08} & 5146.4 \\ 
\multicolumn{1}{l|}{Vicuna-13b} & \multicolumn{1}{c}{3.65} & 3.06 & \multicolumn{1}{c|}{\textbf{3.01}} & \textbf{24.50} & \textbf{23.29} & 3.20 & 3.17 & \multicolumn{1}{c|}{3.37} & 18.96 & 24.93 & \multicolumn{1}{c|}{23.37} & 4182.9 \\
\midrule
\multicolumn{13}{c}{\textit{TunedRR (Retrieve-then-Read on Fine-tuned Models)}} \\ \midrule
\multicolumn{1}{l|}{Llama2-7b-SFT} & \multicolumn{1}{c}{4.06} & \underline{3.78} & \multicolumn{1}{c|}{0.47} & 12.03 & \underline{17.19} & \underline{3.91} & \underline{3.67} & \multicolumn{1}{c|}{\underline{3.34}} & 32.29 & 24.67 & \multicolumn{1}{c|}{21.23} & 740.9 \\
\multicolumn{1}{l|}{Llama2-13b-SFT} & \multicolumn{1}{c}{\textbf{4.22}} & \textbf{3.88} & \multicolumn{1}{c|}{0.17} & 11.39 & \underline{17.19} & \textbf{4.01} & \textbf{3.69} & \multicolumn{1}{c|}{3.32} & \textbf{38.08} & \underline{27.62} & \multicolumn{1}{c|}{\underline{24.85}} & 633.5 \\
\multicolumn{1}{l|}{Vicuna-7b-SFT} & \multicolumn{1}{c}{3.93} & 3.54 & \multicolumn{1}{c|}{\textbf{0.66}} & \textbf{13.45} & 17.08 & 3.57 & 3.03 & \multicolumn{1}{c|}{3.21} & 24.74 & 25.75 & \multicolumn{1}{c|}{22.62} & 1109.6 \\
\multicolumn{1}{l|}{Vicuna-13b-SFT} & \multicolumn{1}{c}{\underline{4.07}} & 3.68 & \multicolumn{1}{c|}{\underline{0.65}} & \underline{12.80} & \textbf{17.26} & \underline{3.91} & \underline{3.67} & \multicolumn{1}{c|}{\textbf{3.40}} & \underline{34.68} & \textbf{29.58} & \multicolumn{1}{c|}{\textbf{26.73}} & 944.6\\
\bottomrule
\end{tabular}
}

\caption{Wikipedia generation results for different combinations of LLMs and generation methods. Cit. stands for citation, MET for METEOR, R-L for ROUGE-L, Org. for Organization, and Cover. for Coverage. The LLM-based scores are \textit{italicized} and range from 0 to 5, while the other metrics range from 0 to 100. The \textbf{best} results for each method are in \textbf{bold}, the \underline{second best} results are \underline{underlined}.}
\label{main-results}
\end{table*}

\subsection{Main Results}
\label{sec:main_results}
In our main experiments, we evaluate three types of frameworks combined with different base LLMs and the default DPR reranker on writing, informativeness, and verifiability. The results are shown in Table \ref{main-results}.

\noindent \textbf{Writing}\quad 
We can see that all methods perform consistently high in terms of fluency. Among the base models we adopt, GPT3.5 stands out as the top performer, achieving impressive \textit{Fluency} and \textit{Organization Scores} that are close to those of human-authored Wikipedia articles. This demonstrates that current LLMs are exceptionally adept at generating organized and readable text, resembling the natural flow and grammatical correctness of human writing. It is also noted that the hierarchical generation methods (PRR) tend to have lower writing performance due to the separate generation of each section, compared to the corresponding RR methods. Benefiting from fine-tuning the output results of GPT4, TunedRR demonstrates improved fluency and coherence. Regarding the outlining ability, we can see that PRR consistently exhibits stable section content planning, regardless of the base model size, compared to RR methods. However, the TunedRR methods perform the worst in outlining, as these models are unable to generate titles in the correct format. Despite this, it does not influence the overall organization score.

\noindent \textbf{Informativeness}\quad
It is evident that the overall information in reference Wikipedia is very rich, as reflected in the high \textit{Info Score}. Among the three methods, weaker base models benefit from the fine-tuning process of TunedRR and exhibit stable performance across different metrics. The overall informativeness of Wikipedia generated using PRR methods tends to be higher than that of RR methods. It is worth noting that even when PRR produces longer Wikipedia articles than reference Wikipedia, there can still be a significant disparity in the richness of information. This may be caused by the excessive amount of content unrelated to the main topic in each section, as revealed by the relatively low \textit{Focus Score} of PRR. Consequently, this may explain why PRR often scores lower than RR in \textit{Coverage Score}. It is also noted that n-gram based metrics like ROUGE-L and METEOR are heavily influenced by length, as pointed out by \citet{krishna-etal-2021-hurdles}. Thus, using multiple metrics is helpful to analyze the performance of the models.

\noindent \textbf{Verifiability}\quad
The base generation model plays a critical role in determining citation capability. In the RR and PRR methods, GPT3.5-based methods outperform others significantly. In contrast, open-source models exhibit much lower citation abilities, with Citation Rate not exceeding 20\%, aligning with~\citet{gao-etal-2023-enabling}. The TunedRR methods demonstrate competitive citation capability. Simple fine-tuning can enhance Citation Recall and Citation Precision by nearly 20\% compared to RR methods. Nevertheless, the upper limit of this fine-tuning is still suboptimal compared to the capabilities of GPT4. In the future, exploring high-quality data for fine-tuning will be crucial to improving verifiability.

\noindent \textbf{Article Length}\quad
Reference Wikipedia articles focusing on recent events have around 1,600 words, while RR methods produce shorter articles, typically around 500 words. Hierarchical PRR methods can generate much longer articles, even over 5,000 words. However, the informativeness of generated articles is not necessarily positively correlated with length but depends on the model's capabilities. For example, GPT3.5 achieves a higher \textit{Info Score}, while the 7B weaker models generate excessively long articles with low \textit{Info Score}.

\subsection{Analysis of Retrieval Setting}
To conduct an in-depth analysis, we explore the retrieval setting, including different reranker techniques, the number of related documents, and citation sources. In these experiments, we use the simple RR method with GPT3.5 as the generator. For the experiment in Table \ref{tab:num_documents}, we use GPT3.5-16k to allow more related documents as input, ensuring a sufficiently long context window.

\begin{table}[t]

\resizebox{\linewidth}{!}{
\begin{tabular}{@{}c|cccccc@{}}
    \toprule
    \multicolumn{1}{c|}{\textbf{\#Docs}} & \textit{\begin{tabular}[c]{@{}c@{}}Fluency \\ Score\end{tabular}} & \textit{\begin{tabular}[c]{@{}c@{}}Org. \\ Score\end{tabular}} & \begin{tabular}[c]{@{}c@{}}R-L \end{tabular} & \begin{tabular}[c]{@{}c@{}}Cit. \\ Recall\end{tabular} & \begin{tabular}[c]{@{}c@{}}Cit. \\ Precision\end{tabular} & \begin{tabular}[c]{@{}c@{}}Length \end{tabular} \\ \midrule
    0 & 4.62 & 4.32 & 16.22 & - & - & 574.7 \\
    5 & 4.29 & 4.02 & 17.42 & 33.98 & 32.68 & 541.3 \\
    10 & 4.30 & 3.99 & 17.80 & \textbf{34.75} & \textbf{32.80} & 559.9 \\
    15 & 4.29 & 4.08 & 18.09 & 32.41 & 30.22 & 583.2 \\
    20 & 4.30 & 4.10 & \textbf{18.44} & 32.85 & 30.85 & 584.9 \\
    \bottomrule
\end{tabular}
}
\caption{Impact of the number of related documents.}
\label{tab:num_documents}
\end{table}

\noindent \textbf{Number of Related Documents}\quad
We experiment with a number of related documents ranging from 0 to 20, as shown in Table \ref{tab:num_documents}. From this table, we see that the length of generated article ranges between 500-600 words and is insensitive to the number of input documents. Without any input, the model can produce a fluent and well-organized article, but none of the generated sentences can be verified.

Overall, the \textit{Fluency} and \textit{Organization Scores} of the model are not significantly affected by the number of related documents. As the number of documents increases, the amount of included information also grows, enhancing the informativeness (ROUGE-L and \textit{Info Score}) of the generated content. At the same time, the verifiability of the model deteriorates with more related documents. The citation quality peaks with around 10 retrieved documents and gradually declines thereafter, indicating that the model may struggle to effectively handle an excessive number of input documents. Therefore, expanding the context window of LLMs may not fully address the challenges of generating full-length Wikipedia articles and could necessitate the integration of more advanced retrieval or reranking methods. The complete experimental results can be found in Table \ref{number:appendix}.

\begin{table}[t]
\resizebox{\linewidth}{!}{
\begin{tabular}{@{}c|cccccc@{}}
\toprule
\multicolumn{1}{c|}{\begin{tabular}[c]{@{}c@{}}\textbf{Retrieval} \\ \textbf{Method}\end{tabular}} & \textit{\begin{tabular}[c]{@{}c@{}}Fluency \\ Score\end{tabular}} & \textit{\begin{tabular}[c]{@{}c@{}}Org. \\ Score\end{tabular}} & \begin{tabular}[c]{@{}c@{}}R-L\end{tabular} & \textit{\begin{tabular}[c]{@{}c@{}}Info \\ Score\end{tabular}} & \begin{tabular}[c]{@{}c@{}}Cit. \\ Recall\end{tabular} & \begin{tabular}[c]{@{}c@{}}Cit. \\ Precision\end{tabular} \\ \midrule
Random & 4.17 & 4.02 & 16.66  & 3.23 & 30.31 & 26.16 \\
TF-IDF & 4.37 & 4.16 & \textbf{18.10} & 3.59 & \textbf{49.31} & \textbf{48.34} \\
BM25 & \textbf{4.39} & 4.19 & 17.38 & \textbf{3.61} & 46.50 & 44.10 \\
DPR & 4.31 & \textbf{4.31} & 17.81 & 3.49 & 42.09 & 36.70 \\
GTR & 4.33 & 4.10 & 17.89  & 3.54 & 44.80 & 40.21 \\ 
\bottomrule
\end{tabular}
}
\caption{Performance of different rerankers with the top 5 documents are used for generation.}
\label{tab:retrievers}
\end{table}

\noindent \textbf{Sparse vs Dense Reranker}
To rerank related documents for generation, we compare widely used sparse rankers (TF-IDF, BM25) and dense rankers (DPR, GTR), as shown in Table \ref{tab:retrievers}. We used a random selection method as the baseline to set clear benchmarks for the worst possible outcomes. From Table \ref{tab:retrievers}, it is evident that articles produced with random reranking performed significantly worse across all metrics. Among all reranking techniques, term-matching sparse rerankers (TF-IDF, BM25) outperform dense retrievers. This aligns with \citet{sciavolino2022simple}, who found that dense retrievers often struggle to identify rare entities not encountered during training, which is a significant issue for Wikipedia. Since sparse rerankers struggle with complex semantic queries and dense rerankers show competitive performance, we choose the commonly adopted DPR method as the default reranker. However, as stated above, LLMs still struggle to effectively utilize all the content within the context length, despite the increasing context length of models. Therefore, reranking techniques are crucial to final performance, and improvements in reranking would benefit the Wikipedia generation task.
\begin{table}

\resizebox{\linewidth}{!}{
\begin{tabular}{@{}c|cccccc@{}}
\toprule
\multicolumn{1}{c|}{\begin{tabular}[c]{@{}c@{}}\textbf{Document} \\ \textbf{Source}\end{tabular}} & \textit{\begin{tabular}[c]{@{}c@{}}Fluency \\ Score\end{tabular}} & \textit{\begin{tabular}[c]{@{}c@{}}Org. \\ Score\end{tabular}} & \begin{tabular}[c]{@{}c@{}}R-L\end{tabular} & \textit{\begin{tabular}[c]{@{}c@{}}Info \\ Score\end{tabular}} & \begin{tabular}[c]{@{}c@{}}Cit. \\ Recall\end{tabular} & \begin{tabular}[c]{@{}c@{}}Cit. \\ Precision\end{tabular} \\ \midrule

Search Engine & \textbf{4.31} & \textbf{4.16} & \textbf{4.33} & \textbf{3.51} & 35.96 & 34.31 \\
Human Editor & 4.31 & 4.05 & 4.28 & 3.43 & 35.71 & 33.76 \\
Mixed & 4.31 & 4.05 & 4.26 & 3.49 & \textbf{38.78} & \textbf{36.7} \\
\bottomrule
\end{tabular}
}

\caption{Impact of different related document source. }
\label{tab:search}
\end{table}

\noindent \textbf{Citation Source}\quad
The related documents come from two sources: search engines and human editors. We analyze how the source of related documents influences Wikipedia article generation. Table \ref{tab:search} presents the generation results using different sources. While it is commonly believed that documents provided by human editors are of higher quality \citep{liu2018generating,qian2023webbrain}, our findings suggest that for new events, search engines can offer more informative and verifiable references. Regarding writing and informativeness, using search engine sources alone performs better than using human editor sources alone or a mix of both. In terms of verifiability, search engine and human editor sources perform similarly, with a mixture performing slightly better than a single source. This indicates that search engines can cover most information provided by human editors. This finding paves the way for the automatic generation of Wikipedia articles for new events, as search engines can provide access to a wide array of up-to-date and extensive news sources, ensuring a breadth and depth of information that rivals or exceeds what human editors can compile.

\subsection{Analysis of Supervised Finetuning}

\begin{figure*}[t]
    \centering
    \includegraphics[width=\textwidth]{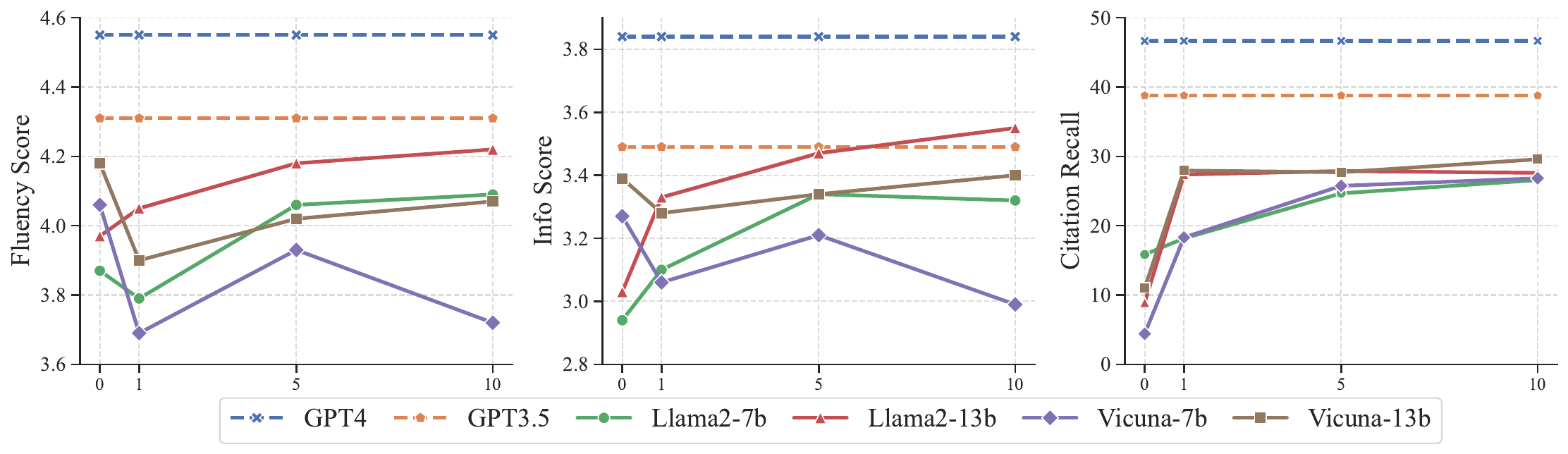} 
    \caption{We fine-tuned models of various scales and families, evaluating checkpoints at 1, 5, and 10 epochs. We selected one primary metric from each of the three dimensions and displayed their performance trends with training epochs.}
    \label{fig:example finetune}
\end{figure*}

This subsection explores how to enhance performance during tuning. We selected one representative metric from each of the three dimensions and plotted their performance trends with training epochs in Figure \ref{fig:example finetune}. In the Writing dimension, initial tuning rounds show a decline in fluency, but subsequent epochs reveal improved writing techniques, resulting in progressively higher \textit{Fluency Scores}. After ten epochs, most models exceed their original performance, except for Vicuna-13b, likely due to its initially strong writing abilities. For the Informativeness dimension, a similar trend is observed: information richness initially declines but then recovers and surpasses initial performance levels. Llama2-13b even surpasses GPT3.5 in terms of \textit{Info Score}. However, Vicuna-7b shows good performance by the fifth epoch but overfits after ten epochs, leading to a decline. In the Verifiability dimension, even one training epoch significantly enhances citation abilities. Further training improves citation accuracy, though the improvement rate slows after five epochs. Despite supervised training, open-source models still lag behind proprietary models in performance.

\section{Related Work}
\subsection{Retrieval-Augmented Text Generation}
Enhancing LLMs with retrieval mechanisms during inference has become a common practice for generative tasks~\citep{li2022survey, gao2023retrieval, guu2020retrieval}. In the era of LLMs, RAG has proven to be an effective and versatile paradigm across various NLP tasks \citep{weston2018retrieve, jiang2023active}. Studies have demonstrated that retrieval can provide in-context examples \citep{brown2020language}, thereby enhancing the capabilities of LLMs \citep{huang2023raven, ram2023context, chen2023exploring}. The concept of RAG was first introduced by \citet{lewis2020retrieval}, who combined a pre-trained seq2seq model with a non-parametric dense vector index of Wikipedia, setting new state-of-the-art benchmarks for open-domain question answering tasks. To address the tendency of LLMs to produce hallucinations, researchers \citep{nakano2021webgpt} have proposed integrating language models with result pages from search engines to refine the final output. Systems may perform multiple retrieval processes during generation,  combined with techniques such as Chain of Thought \citep{trivedi2022interleaving, feng2023retrieval}.

\subsection{Automated Wikipedia Generation}
The task of automatically generating Wikipedia articles has garnered significant research attention over the past decade~\cite{sauper2009automatically,liu2018generating,perez2019generating,banerjee2016wikiwrite}. \citet{sauper2009automatically} introduced a structure-aware method for synthesizing Wikipedia articles from relevant source documents. \citet{banerjee2016wikiwrite} made further advancements with the development of WikiWrite, which classified retrieved information by analyzing relationships between referenced and target entities.

The emergence of pretrained language models (PLMs) marked a paradigm shift in Wikipedia generation, enhancing both the structuring of documents and ensuring factual accuracy. \citet{liu2018generating} demonstrated a significant breakthrough by employing a multi-document summarization strategy with a decoder-only transformer model, effectively leveraging PLMs for generating coherent text. Building upon this foundation, \citet{fan2022generating} introduced a retrieval mechanism to extract supporting information from the web. They utilized BART to generate long-form biographies section by section, guided by predefined headings. Their work emphasizes both sectioning and citations. However, the evaluation process remained largely dependent on manual assessment. \citet{qian2023webbrain} proposed WebBrain, a sophisticated system designed to produce short factual articles using a web corpus. WebBrain curated a larger dataset comprising the leading sections of Wikipedia articles. The generated articles required citations to ensure verifiability, and the evaluation employed methods such as QG-QA~\citep{goodrich2019assessing,wang2020asking} to assess factual correctness. 
\citet{shao2024assisting} refines the generation of section headings through iterative questioning with LLM-based metrics for evaluation. However, their evaluation was conducted on only 100 samples, making it difficult to fully validate the new method's overall performance.

\section{Conclusion}
In this paper, we introduce \bench{} to address the challenge of generating full-length Wikipedia articles under real-world scenarios. We have introduced advanced metrics to systematically evaluate the performance of Wikipedia generation. Our experiments with three RAG frameworks demonstrate the potential of LLMs to generate coherent and informative Wikipedia articles. We explore various retrieval settings and examine the impact of different citation sources. We highlight the importance of high-quality data for fine-tuning to improve article verifiability. Overall, this work compares LLM-based methodologies for full-length Wikipedia generation, providing insights and guiding future research.

\section*{Limitations}
Our research encounters limitations, notably in our section-by-section generation approach, which may lead to redundancy and necessitate a rewriting strategy to ensure article cohesion. Further limitations include the challenge of direct citation by LLMs, a bottleneck that might necessitate the exploration of post-citation methods such as employing NLI for improvement. Additionally, the information from related documents does not fully cover the content of original Wikipedia articles, making the n-gram metric comparison between the generated text and the original articles a weak reference rather than a definitive standard. 


\section*{Ethics Statement}

Unlike creative content generation, grounded article generation can influence how people learn about topics and consume source information. To mitigate potential harm, we implement strict verifiability checks to ensure the reliability of the output. 

Additionally, our information sources primarily come from the internet, which inevitably includes inaccurate or unreliable content. While we can provide the origin of citations, additional effort is required to verify the reliability of the information itself. Improved retrieval-augmented generation (RAG) methods, along with sourcing from more authoritative websites, may help mitigate this issue.

Another ethical limitation of this work is that we have focused solely on generating English Wikipedia articles. Expanding the system to support multilingual article generation is a valuable direction for future research, as many topics lack coverage in non-English Wikipedia pages.

\section*{Acknowledgement}
We thank anonymous reviewers for their helpful comments on this paper. This work was partially supported by National Natural Science Foundation of China (No. 92470205, 62476010). Sujian Li is the corresponding author. 

\bibliography{anthology,custom,acl2023}

\appendix
\newpage
\appendix
\onecolumn
\section{Evaluation Metrics Comparison}
\label{appendix:evaluation}

\begin{table}[htbp]
\centering
\setlength{\tabcolsep}{4pt} 
\renewcommand{\arraystretch}{1.2} 
\resizebox{0.9\textwidth}{!}{ 
\begin{tabular}{@{}c|cccc@{}}
\toprule
\textbf{Relevant Work}                        & \textbf{Writing}                                                                                                                                                        & \textbf{Informativeness}                                                                                                     & \textbf{Faithfulness/Verifiability}                        & \textbf{Full Length} \\ \midrule
\cite{banerjee2016wikiwrite} & \textcolor{red}{\XSolidBrush}                                                                                                                                                                       & ROUGE                                                                                                                        & \textcolor{red}{\XSolidBrush}                                                          & \textcolor{green}{\CheckmarkBold}                    \\
\midrule[0.5pt]
\cite{liu2018generating}     & \HandPencilLeft                                                                                                                                                                       & \begin{tabular}[c]{@{}c@{}}ROUGE\\ log-perplexity\end{tabular}                                                               & \textcolor{red}{\XSolidBrush}                                                           & \textcolor{green}{\CheckmarkBold}                    \\
\midrule[0.5pt]
\cite{perez2019generating}   & \HandPencilLeft                                                                                                                                                                       & \begin{tabular}[c]{@{}c@{}}ROUGE\\ Unigram F-measure\end{tabular}                                                            & \textcolor{red}{\XSolidBrush}                                                          & \textcolor{red}{\XSolidBrush}                    \\
\midrule[0.5pt]
\cite{liu2019hierarchical}   & \HandPencilLeft                                                                                                                                                                       & ROUGE                                                                                                                        & \textcolor{red}{\XSolidBrush}                                                          & \textcolor{red}{\XSolidBrush}                     \\
\midrule[0.5pt]
\cite{logan2021fruit}        & \textcolor{red}{\XSolidBrush}                                                                                                                                                                       & UpdateROUGE                                                                                                                  & Entity Prec. \& Recall                                     & \textcolor{red}{\XSolidBrush}                     \\
\midrule[0.5pt]
\cite{fan2022generating}     & \textcolor{red}{\XSolidBrush}                                                                                                                                                                       & \begin{tabular}[c]{@{}c@{}}ROUGE\\ NER Coverage\end{tabular}                                                                 & NLI                                                        & \textcolor{green}{\CheckmarkBold}                    \\
\midrule[0.5pt]
\cite{qian2023webbrain}      & \HandPencilLeft                                                                                                                                                                       & \begin{tabular}[c]{@{}c@{}}BLEU\\ METEOR\\ ROUGE\\ CIDEr\end{tabular}                                                        & \begin{tabular}[c]{@{}c@{}}QAGS\\ TripleScore\end{tabular} & \textcolor{red}{\XSolidBrush}                    \\
\midrule[0.5pt]
\cite{shao2024assisting}     & \begin{tabular}[c]{@{}c@{}}Organization Score by Prometheus\\ Focus Score by Prometheus \\ Outline Recall by Prometheus\end{tabular}                                            & \begin{tabular}[c]{@{}c@{}}ROUGE\\ Entity Recall\\ Coverage by Prometheus\end{tabular}                                       & NLI(Citation quality)                                      & \textcolor{green}{\CheckmarkBold}                    \\ \midrule
\textbf{Ours}                                 & \begin{tabular}[c]{@{}c@{}}Fluency Score by GPT-4 \\  Outline Score by GPT-4\\ Organization Score by Prometheus\\ Relevance Score by Prometheus \end{tabular} & \begin{tabular}[c]{@{}c@{}} BLEU\\ METEOR\\ ROUGE\\ Info Score by GPT-4\\ Coverage Score by Prometheus \\ Focus Score by Prometheus \end{tabular} & NLI(Citation quality)                                      & \textcolor{green}{\CheckmarkBold}                    \\ \bottomrule
\end{tabular}
}
\caption{Comparison of Evaluation Metrics Across Different Works. \HandPencilLeft \ indicates human evaluation, \textcolor{red}{\XSolidBrush} indicates no evaluation or the work is about short Wikipedia snippet, and \textcolor{green}{\CheckmarkBold} signifies the work is about full-length Wikipedia generation. Metrics include writing quality, informativeness, and verifiability.}
\end{table}

\newpage
\section{\textbf{WikiGenBen} Datasest}
\label{wikipages}

\subsection{Wikipedia Reference}
This section provides a detailed example of a data entry for the "2023 USFL season," illustrating the dataset's structure and information richness.
Our dataset includes detailed information of Wikipedia reference articles, acquired using the Python library MediaWiki.
\begin{tcolorbox}
\begin{itemize}
    \item \textbf{ID:} 71284256
    \item \textbf{Keyword:} 2023 USFL season
    \item \textbf{URL:} \url{https://en.wikipedia.org/wiki/2023%20USFL%20season}
    \item \textbf{Summary:} The 2023 USFL season was the second season of the United States Football League. The regular season started on April 15 and ended on June 18. The postseason began on June 24 and ended with the 2023 USFL Championship Game on July 1. The league expanded the locations their teams play to four total stadiums, adding Ford Field in Detroit, Michigan, and Simmons Bank Liberty Stadium in Memphis, Tennessee.
    \item \textbf{Sections:} Offseason, Locations, Teams, Players, ...
    \item \textbf{Content for each section:} 
    \begin{enumerate}
        \item During the 2022 season, …
        \item The league stated its …
        \item On November 15, 2022, in conjunction…
        \item For the 2023 season, each USFL team…
        \item ...
    \end{enumerate}
    \item \textbf{Infobox:}  \\
    \\
    \begin{tabular}{|l|p{10cm}|}
    \hline
    \textbf{Key} & \textbf{Value} \\
    \hline
    League & United States Football League \\
    Sport & American football \\
    Duration & Regular season: April 15 – June 18 Playoffs: June 24 – July 1 \\
    ... & ...\\
    \hline
    \end{tabular}
\end{itemize}
\end{tcolorbox}

\subsection{Related Documents}
Beyond the core Wikipedia entries, our dataset includes related documents categorized as 'Human' and 'Search'. 'Human' documents come from the Wikipedia External Links Section, offering human-curated, credible information. 'Search' documents are obtained through Google searches, providing diverse perspectives and additional context.
\begin{tcolorbox}
\begin{itemize}
    \item \textbf{Doc ID:} 1
    \item \textbf{Title:} Johnson, Roy S. (2022-11-14). \"USFL reveals season 2 details for Birmingham\". al. Retrieved 2022-12-13.
    \item \textbf{URL:} https://www.al.com/news/2022/11/usfl-reveals-season-2-details-for-birmingham.html
    \item \textbf{Content:} usfl reveals season 2 details for...
    \item \textbf{Source:} Human / Search
\end{itemize}
\end{tcolorbox}

\subsection{Training Data} \label{appendix:training}
Wikipedia text is typically used for training. However, due to the high failure rate of Wikipedia's related document links, we use an alternative method. We employ DPR reranking to fetch the top-5 documents for each entry and then use the RR prompt to generate text with GPT-4. This approach aims to teach the target model GPT-4's citation generation capabilities.

\section{Prompt for Generating} \label{Prompt design}
Our method involves creating a base template for the prompt, which is then supplemented with relevant documents until reaching a maximum input length. Specifically, for 4k models, the maximum input length is strictly capped at 2048 tokens.
\subsection{Retrieve-then-Read} \label{RR prompt}
The \textbf{RR} approach involves a straightforward, one-stage process for directly generating an article. 
\begin{tcolorbox}[title = {Article Generation Prompt}]

\textbf{Input}:\\
I have a topic "\{\textbf{keyword}\}" that contains the following documents: \\
Document 1: \{\textbf{doc1}\} \\
Document 2: \{\textbf{doc2}\} \\
... \\
Based on the above information, you are assigned to write a Wikipedia article on the topic. \\
Organize the content of your article by sections. Before writing each section, always starts with "==SECTION NAME==". \\
You must cite the most relevant document for every sentence you write, in the format of "This is an example sentence.[k]", where k denotes Document k. \\

\end{tcolorbox}

\subsection{Plan-Retrieve-Read}
In contrast, the \textbf{PRR} method necessitates a more structured approach to article generation. Initially, it requires the planning of an article outline. Once the outline is established, each section name generated during the planning phase serves as a guide for the subsequent retrieval and writing phases.
\begin{tcolorbox}[title = {Outline Generation Prompt}]

\textbf{Input}:\\
I have a topic "\{\textbf{keyword}\}" that contains the following documents: \\
Document 1: \{\textbf{doc1}\} \\
Document 2: \{\textbf{doc2}\} \\
... \\
Based on the above information, you are assigned to write an outline for a Wikipedia article about this topic. \\
Your outline should only include the names of the sections, without any further details. \\
Do not use document name as your outline. \\
The format of your outline should be as follows: \\
1. Introduction \\
2. <Section Name 1> \\
... \\
n. <Section Name n> \\
\end{tcolorbox}

\begin{tcolorbox}[title = {Section Generation Prompt}]

\textbf{Input}:\\
I have a topic "\{\textbf{keyword}\}" and a section "\{\textbf{section}\}" that contains the following documents: \\
Document 1: \{\textbf{doc1}\} \\
Document 2: \{\textbf{doc2}\} \\
... \\
Based on the above information, you are assigned to write a Wikipedia article on the topic. \\
You must cite the most relevant document for every sentence you write, in the format of "This is an example sentence.[k]", where k denotes Document k. \\
\end{tcolorbox}

\section{Prompt for Evaluating}
\label{evaluation-appendix}
We employed the GPT-4-1106-preview model by OpenAI for scoring, setting the temperature to 0 and keeping other parameters at their defaults. Regular expressions were used to match the corresponding scores. As LLM-EVAL\citep{lin-chen-2023-llm} shows that a single prompt can obtain multi-dimensional scores correlating well with human preferences, we called the GPT-4 API only once to get the Fluency and Informativeness Scores. This approach significantly reduces costs by eliminating the need for multiple prompts.
\begin{tcolorbox}[title = {Evaluation Prompt for Fluent Score and Informativeness Score}]

\textbf{Input}: \\
Evaluate an encyclopedia text of a keyword on three metrics: fluency, informativeness, and faithfulness.

Give a score from 0-5 for each metric. 

- Fluency: Assess the text for grammatical correctness, coherence of ideas, and overall readability. Look for smooth transitions between sentences and paragraphs, as well as clear organization of information.

- Informativeness: Evaluate the depth and breadth of information provided about the keyword. Check if the text covers various aspects of the topic, including its definition, background, significance, related concepts, and any relevant examples or applications.

- Faithfulness: Verify the accuracy of the information presented in the text by cross-referencing with credible sources or established knowledge. Assess whether the information aligns with accepted facts and evidence.\\\\

Only give three scores in the form of: Fluency: Score 1, Informativeness: Score 2, Faithfulness: Score 3. No need for explanation.
\end{tcolorbox}
The GPT-4-1106-preview model is trained on data up to April 2023. Since nearly one-third of events in our benchmark occurred after April 2023, GPT-4-1106-preview cannot evaluate faithfulness accurately. Therefore, we do not report the faithfulness score.
\begin{tcolorbox}[title = {Evaluation Prompt for Outline Score}]

\textbf{Input}:

Given a keyword and an outline about the Wikipedia of the keyword, assign a score ranging from 0 to 5 to evaluate the quality of the outline. Only give the score without explanation.
\end{tcolorbox}

We adopt the same scoring rubrics as previous studies~\citep{shao2024assisting} to assess Organization, Focus and Coverage, with the exception of replacing Prometheus with Prometheus-2 for an extended context length of 32k.
\begin{tcolorbox}[title = {Criteria Description for Organization Score, Focus Score and Coverage Score}]

\textbf{ Coherence and Organization:} 

Is the article well-organized and logically structured? 

- Score 1: Disorganized; lacks logical structure and coherence. 

- Score 2: Fairly organized; a basic structure is present but not consistently followed. 

- Score 3: Organized; a clear structure is mostly followed with some lapses in coherence. 

- Score 4: Good organization; a clear structure with minor lapses in coherence. 

- Score 5: Excellently organized; the article is logically structured with seamless transitions and a clear argument. \\\\

\textbf{Relevance and Focus:} 

Does the article stay on topic and maintain a clear focus? 

- Score 1: Off-topic; the content does not align with the headline or core subject. 

- Score 2: Somewhat on topic but with several digressions; the core subject is evident but not consistently adhered to. 

- Score 3: Generally on topic, despite a few unrelated details. 

- Score 4: Mostly on topic and focused; the narrative has a consistent relevance to the core subject with infrequent digressions. 

- Score 5: Exceptionally focused and entirely on topic; the article is tightly centered on the subject, with every piece of information contributing to a comprehensive understanding of the topic.\\\\ 

\textbf{Broad Coverage:} 

Does the article provide an in-depth exploration of the topic and have good coverage? 

- Score 1: Severely lacking; offers little to no coverage of the topic’s primary aspects, resulting in a very narrow perspective. 

- Score 2: Partial coverage; includes some of the topic’s main aspects but misses others, resulting in an incomplete portrayal. 

- Score 3: Acceptable breadth; covers most main aspects, though it may stray into minor unnecessary details or overlook some relevant points. 

- Score 4: Good coverage; achieves broad coverage of the topic, hitting on all major points with minimal extraneous information.

- Score 5: Exemplary in breadth; delivers outstanding coverage, thoroughly detailing all crucial aspects of the topic without including irrelevant information.

\end{tcolorbox}

\newpage

\section{Generation Format}

For generation, we require the model to append references in the format [1][2] after each claim, indicating support by chunk 1 and chunk 2. Additionally, we generate section titles using '==' as delimiters. We decompose the article accordingly and evaluate the quality of citations and section titles.

\begin{tcolorbox}[title = {An example of model output}]

\textbf{2022 EFL League One play-off final}:

\begin{center}==Introduction==\end{center}
The 2022 EFL League One play-off final is a highly anticipated football match that will determine the final promotion spot to the EFL Championship for the 2022-2023 season. The play-off final is the culmination of the EFL League One play-offs, which involve the teams that finish in the third to sixth positions in the league table. The winner of the final will join the top two teams in the league in gaining promotion to the higher division. This article will provide an overview of the 2022 EFL League One play-off final, including the date, venue, and teams involved.

\begin{center}==Date and Venue==\end{center}
The 2022 EFL League One play-off final is scheduled to take place on Saturday, May 21, 2022.[2] The final will be held at Wembley Stadium in London, England, which has been the traditional venue for the EFL play-off finals.[2] Wembley Stadium is known for its iconic status in English football and has a seating capacity of over 90,000 spectators.[2]

\begin{center}==Teams==\end{center}
The teams that will compete in the 2022 EFL League One play-off final will be determined through the League One play-off semi-finals. The semi-finals will involve the teams that finish in the third to sixth positions in the League One table.[2] The semi-final matches are scheduled to take place on Thursday, May 5, 2022, and Friday, May 6, 2022.[2] The winners of the semi-finals will advance to the final and compete for the promotion spot to the EFL Championship.

\begin{center}==Ticket Information==\end{center}
Tickets for the 2022 EFL League One play-off final are expected to go on sale in May.[3] The English Football League (EFL) will provide more information on ticket sales closer to the date of the final.[3] In previous years, the play-off finals have attracted a significant number of spectators, with Wembley Stadium hosting capacity-limited crowds of just under 12,000.[3] However, this year, with the return to full capacity crowds, the atmosphere is expected to be even more electric.

\begin{center}==Broadcasting==\end{center}
The 2022 EFL League One play-off final, along with the semi-finals, will be televised by Sky Sports, allowing fans to watch the matches from the comfort of their homes.[3] Additionally, highlights of the finals will be available on Quest, providing fans with a recap of the key moments from the matches.[3]

\begin{center}==Conclusion==\end{center}
The 2022 EFL League One play-off final is set to be an exciting and highly anticipated event in English football. Taking place at Wembley Stadium, the final will determine the final promotion spot to the EFL Championship for the 2022-2023 season. With tickets expected to go on sale in May and the matches being broadcasted on Sky Sports, fans will have the opportunity to witness the drama and excitement of the play-off final.

\end{tcolorbox}

\newpage
\section{Impact of Retrieval Settings}

\begin{table}[h]
\resizebox{\columnwidth}{!}{%
\begin{tabular}{c|ccc|ccccc|ccc|c}
\hline
\multirow{2}{*}{\textbf{Doc \#}} & \multicolumn{3}{c|}{\textbf{Writing}} & \multicolumn{5}{c|}{\textbf{Informativeness}} & \multicolumn{3}{c|}{\textbf{Verifiability}} & \multirow{2}{*}{\textbf{Length}} \\
 & \textit{\begin{tabular}[c]{@{}c@{}}Fluency\\ Score\end{tabular}} & \begin{tabular}[c]{@{}c@{}}Org.\\ Score\end{tabular} & \begin{tabular}[c]{@{}c@{}}Outline\\ Score\end{tabular} & MET & R-L & \textit{\begin{tabular}[c]{@{}c@{}}Focus\\ Score\end{tabular}} & \textit{\begin{tabular}[c]{@{}c@{}}Cover.\\ Score\end{tabular}} & \textit{\begin{tabular}[c]{@{}c@{}}Info\\ Score\end{tabular}} & \textit{\begin{tabular}[c]{@{}c@{}}Cit. \\ Rate\end{tabular}} & \begin{tabular}[c]{@{}c@{}}Cit. \\ Recall\end{tabular} & \begin{tabular}[c]{@{}c@{}}Cit. \\ Prec.\end{tabular} &  \\ \hline
0 & \textbf{4.62} & \textbf{4.32} & \textbf{2.91} & 10.51 & 16.22 & \textbf{4.61} & \textbf{4.29} & \textbf{3.70} & - & - & - & 574.7 \\
5 & 4.29 & 4.02 & 2.84 & 10.29 & 17.42 & 4.22 & 3.89 & 3.39 & \textbf{38.38} & 33.98 & 32.68 & 541.3 \\
10 & 4.30 & 3.99 & 2.80 & 10.54 & 17.80 & 4.18 & 3.83 & 3.44 & 37.70 & \textbf{34.75} & \textbf{32.80} & 559.9 \\
15 & 4.29 & 4.08 & 2.83 & 10.84 & 18.09 & 4.28 & 3.94 & 3.52 & 35.47 & 32.41 & 30.22 & 583.2 \\
20 & 4.30 & 4.10 & 2.82 & \textbf{10.99} & \textbf{18.44} & 4.33 & 3.83 & 3.55 & 34.80 & 32.85 & 30.85 & 584.9 \\ \hline
\end{tabular}%
}
\caption{Impact of the Number of Retrieved Documents}
\label{number:appendix}
\end{table}

\begin{table}[h]
\resizebox{\columnwidth}{!}{%
\begin{tabular}{c|ccc|ccccc|ccc|c}
\hline
\multirow{2}{*}{\textbf{\begin{tabular}[c]{@{}c@{}}Rerank\\ Method\end{tabular}}} & \multicolumn{3}{c|}{\textbf{Writing}} & \multicolumn{5}{c|}{\textbf{Informativeness}} & \multicolumn{3}{c|}{\textbf{Verifiability}} & \multirow{2}{*}{\textbf{Length}} \\
 & \textit{\begin{tabular}[c]{@{}c@{}}Fluency\\ Score\end{tabular}} & \begin{tabular}[c]{@{}c@{}}Org.\\ Score\end{tabular} & \begin{tabular}[c]{@{}c@{}}Outline\\ Score\end{tabular} & MET & R-L & \textit{\begin{tabular}[c]{@{}c@{}}Focus\\ Score\end{tabular}} & \textit{\begin{tabular}[c]{@{}c@{}}Cover.\\ Score\end{tabular}} & \textit{\begin{tabular}[c]{@{}c@{}}Info\\ Score\end{tabular}} & \textit{\begin{tabular}[c]{@{}c@{}}Cit. \\ Rate\end{tabular}} & \begin{tabular}[c]{@{}c@{}}Cit. \\ Recall\end{tabular} & \begin{tabular}[c]{@{}c@{}}Cit. \\ Prec.\end{tabular} &  \\ \hline
Random & 4.17 & 4.02 & 2.80 & 9.99 & 16.66 & 4.28 & 3.71 & 3.23 & 30.31 & 27.26 & 26.16 & 534.0 \\
TF-IDF & 4.37 & 4.16 & \textbf{2.89} & 10.67 & \textbf{18.10} & \textbf{4.36} & \textbf{4.04} & 3.59 & \textbf{49.31} & \textbf{50.32} & \textbf{48.34} & 576.9 \\
BM25 & \textbf{4.39} & 4.19 & 2.83 & 10.17 & 17.38 & \textbf{4.36} & 3.95 & \textbf{3.61} & 46.50 & 46.74 & 44.10 & 546.8 \\
DPR & 4.31 & \textbf{4.31} & 2.86 & \textbf{10.73} & 17.81 & 3.49 & 3.57 & 3.49 & 42.09 & 38.78 & 36.70 & 577.2 \\
GTR & 4.33 & 4.10 & 2.86 & 10.46 & 17.89 & 4.31 & 3.97 & 3.54 & 44.80 & 41.97 & 40.21 & 560.0 \\ \hline
\end{tabular}%
}
\caption{Performance of different rerankers.}
\label{retrievers:appendix}
\end{table}

\begin{table}[h]
\resizebox{\columnwidth}{!}{%
\begin{tabular}{c|ccc|ccccc|ccc|c}
\hline
\multirow{2}{*}{\textbf{\begin{tabular}[c]{@{}c@{}}Information\\ Source\end{tabular}}} & \multicolumn{3}{c|}{\textbf{Writing}} & \multicolumn{5}{c|}{\textbf{Informativeness}} & \multicolumn{3}{c|}{\textbf{Verifiability}} & \multirow{2}{*}{\textbf{Length}} \\
 & \textit{\begin{tabular}[c]{@{}c@{}}Fluency\\ Score\end{tabular}} & \begin{tabular}[c]{@{}c@{}}Org.\\ Score\end{tabular} & \begin{tabular}[c]{@{}c@{}}Outline\\ Score\end{tabular} & MET & R-L & \textit{\begin{tabular}[c]{@{}c@{}}Focus\\ Score\end{tabular}} & \textit{\begin{tabular}[c]{@{}c@{}}Cover.\\ Score\end{tabular}} & \textit{\begin{tabular}[c]{@{}c@{}}Info\\ Score\end{tabular}} & \textit{\begin{tabular}[c]{@{}c@{}}Cit. \\ Rate\end{tabular}} & \begin{tabular}[c]{@{}c@{}}Cit. \\ Recall\end{tabular} & \begin{tabular}[c]{@{}c@{}}Cit. \\ Prec.\end{tabular} &  \\ \hline
Search Engine & \textbf{4.31} & \textbf{4.16} & 2.84 & 10.23 & 17.54 & \textbf{4.33} & 3.86 & \textbf{3.51} & 40.37 & 35.96 & 34.31 & 542.8 \\
Human Editor & \textbf{4.31} & 4.05 & 2.77 & 10.36 & 17.70 & 4.28 & 3.92 & 3.43 & 38.70 & 35.71 & 33.76 & 540.8 \\
Mixed & \textbf{4.31} & 4.05 & \textbf{2.86} & \textbf{10.79} & \textbf{17.89} & 4.26 & \textbf{3.94} & 3.49 & \textbf{42.09} & \textbf{38.78} & \textbf{36.70} & 579.1 \\ \hline
\end{tabular}%
}
\caption{Influence of different related document source.}
\label{source:appendix}
\end{table}

\newpage
\section{Full Evaluation Results on Finetuned Models}

\begin{table*}[h]
\renewcommand{\arraystretch}{1.1}
\centering
\resizebox{\linewidth}{!}{
\begin{tabular}{l|c|ccc|ccccc|ccc|c}
\hline
\multicolumn{1}{c|}{\multirow{2}{*}{\textbf{\begin{tabular}[c]{@{}c@{}}Rerank\\ Methods\end{tabular}}}} & \multirow{2}{*}{\textbf{\begin{tabular}[c]{@{}c@{}}Training\\ Epochs\end{tabular}}} & \multicolumn{3}{c|}{\textbf{Writing}} & \multicolumn{5}{c|}{\textbf{Informativeness}} & \multicolumn{3}{c|}{\textbf{Verifiability}} & \multirow{2}{*}{\textbf{Length}} \\
\multicolumn{1}{c|}{} &  & \textit{\begin{tabular}[c]{@{}c@{}}Fluency\\ Score\end{tabular}} & \textit{\begin{tabular}[c]{@{}c@{}}Org.\\ Score\end{tabular}} & \textit{\begin{tabular}[c]{@{}c@{}}Outline\\ Score\end{tabular}} & \textit{MET} & \textit{R-L} & \textit{\begin{tabular}[c]{@{}c@{}}Focus\\ Score\end{tabular}} & \textit{\begin{tabular}[c]{@{}c@{}}Cover.\\ Score\end{tabular}} & \textit{\begin{tabular}[c]{@{}c@{}}Info\\ Score\end{tabular}} & \textit{\begin{tabular}[c]{@{}c@{}}Cit. \\ Rate\end{tabular}} & \textit{\begin{tabular}[c]{@{}c@{}}Cit. \\ Recall\end{tabular}} & \textit{\textbf{\begin{tabular}[c]{@{}c@{}}Cit. \\ Prec.\end{tabular}}} &  \\ \hline
\multicolumn{1}{c|}{GPT4} & \textbf{-} & 4.55 & 4.40 & 2.88 & 10.74 & 18.08 & 4.57 & 4.29 & 3.84 & 54.27 & 46.65 & 40.84 & 569.0 \\ \hline
\multicolumn{1}{c|}{Llama2-7B} & 0 & 3.87 & 3.64 & 1.43 & 10.21 & 16.05 & 3.77 & 3.27 & 2.94 & 10.16 & 15.85 & 15.83 & 625.7 \\
 & 1 & 3.79 & 3.27 & 0.48 & 12.58 & 16.08 & 3.78 & 3.41 & 3.10 & 30.04 & 18.17 & 15.01 & 1113.1 \\
 & 5 & 4.06 & 3.78 & 0.47 & 12.03 & 17.19 & 3.91 & 3.67 & 3.34 & 32.29 & 24.67 & 21.23 & 740.9 \\
 & 10 & 4.09 & 3.76 & 0.39 & 11.73 & 17.23 & 3.93 & 3.74 & 3.32 & 33.38 & 26.58 & 23.20 & 699.4 \\ \hline
\multicolumn{1}{c|}{Llama2-13B} & 0 & 3.97 & 4.16 & 2.39 & 9.74 & 15.89 & 4.38 & 3.91 & 3.03 & 7.91 & 8.91 & 8.91 & 552.9 \\
 & 1 & 4.05 & 3.52 & 0.09 & 10.58 & 16.70 & 3.66 & 3.39 & 3.33 & 37.07 & 27.38 & 25.73 & 672.3 \\
 & 5 & 4.18 & 3.80 & 0.17 & 11.30 & 17.26 & 3.98 & 3.67 & 3.47 & 38.68 & 27.88 & 24.29 & 643.3 \\
 & 10 & 4.22 & 3.88 & 0.17 & 11.39 & 17.19 & 4.01 & 3.69 & 3.55 & 38.08 & 27.62 & 24.85 & 633.5 \\ \hline
\multicolumn{1}{c|}{Vicuna-7B} & 0 & 4.06 & 3.46 & 1.61 & 10.18 & 17.34 & 3.69 & 3.39 & 3.27 & 6.40 & 4.41 & 4.38 & 535.2 \\
 & 1 & 3.69 & 3.27 & 0.26 & 12.11 & 16.26 & 3.57 & 3.03 & 3.06 & 25.89 & 18.30 & 17.11 & 1006.0 \\
 & 5 & 3.93 & 3.54 & 0.66 & 13.45 & 17.08 & 3.78 & 3.44 & 3.21 & 24.74 & 25.75 & 22.62 & 1109.6 \\
 & 10 & 3.72 & 3.19 & 0.39 & 15.27 & 16.78 & 3.49 & 3.25 & 2.99 & 18.03 & 26.84 & 23.85 & 1444.9 \\ \hline
\multicolumn{1}{c|}{Vicuna-13B} & 0 & 4.18 & 3.72 & 2.27 & 9.80 & 17.33 & 3.98 & 3.63 & 3.39 & 16.88 & 11.03 & 10.70 & 491.8 \\
 & 1 & 3.90 & 3.49 & 0.16 & 11.33 & 16.51 & 3.74 & 3.46 & 3.28 & 36.96 & 27.95 & 26.29 & 814.0 \\
 & 5 & 4.02 & 3.68 & 0.77 & 12.92 & 17.36 & 4.01 & 3.60 & 3.34 & 30.68 & 27.71 & 25.25 & 984.3 \\
 & 10 & 4.07 & 3.68 & 0.65 & 12.80 & 17.26 & 3.91 & 3.67 & 3.40 & 34.68 & 29.58 & 26.73 & 944.6 \\ \hline
\end{tabular}}
\caption{We selected different model checkpoints during training and evaluated their performance on testset.}
\end{table*}

\label{sec:appendix}

\end{document}